\documentclass[11pt,a4paper]{article}
\usepackage[hyperref]{acl2020}
\usepackage{times}
\usepackage{latexsym}

\usepackage{microtype}
\usepackage{graphicx}
\usepackage{amsfonts}
\usepackage{amsmath}
\usepackage{amssymb}
\usepackage{amsthm}
\usepackage{booktabs}

\newcommand{\B}[1]{\mathbf{#1}}
\newcommand{\T}[1]{\mathtt{#1}}
\newcommand{\tb}[1]{\textbf{#1}}
\newcommand{\ti}[1]{\textit{#1}}
\newcommand{\Cal}[1]{\mathcal{#1}}
\newcommand{\Sref}[1]{\S\ref{#1}}

\aclfinalcopy 

\title{Exploring Controllable Text Generation Techniques}

\author{Shrimai Prabhumoye, Alan W Black, Ruslan Salakhutdinov \\
  School of Computer Science \\
  Carnegie Mellon University \\
  Pittsburgh, PA, USA \\
  \texttt{\{sprabhum, awb, rsalakhu@cs.cmu.edu\}} \\}

\date{}

\begin{document}
\maketitle
\begin{abstract}
Neural controllable text generation is an important area gaining attention due to its plethora of applications.
Although there is a large body of prior work in controllable text generation, there is no unifying theme.
In this work, we provide a new schema of the pipeline of the generation process by classifying it into five modules.
The control of attributes in the generation process requires modification of these modules.
We present an overview of different techniques used to perform the modulation of these modules.
We also provide an analysis on the advantages and disadvantages of these techniques.
We further pave ways to develop new architectures based on the combination of the modules described in this paper.
\end{abstract}

\section{Introduction}
\label{intro}
Controllable text generation is the task of generating natural sentences whose attributes can be controlled.  
The attributes to control can range from being stylistic such politeness, sentiment, formality, etc.; demographic attributes of the person writing the text such as gender, age, etc.; content such as information, keywords, entities, etc.; ordering of information, events, like plot summaries etc. 
Controlling various attributes of text generation has manifold applications.
For instance in dialogue response generation task, work has been done in controlling persona \cite{zhang2018personalizing,li2016persona}, controlling various aspects of the response such as politeness \cite{niu:2018}, formality, authority etc, grounding the responses in external source of information \cite{zhou2018dataset,dinan2018wizard,ghazvininejad2018knowledge}, and controlling topic sequence \cite{tang2019target,prabhumoye2020i}.
Another application is story generation where you can control the ending \cite{peng2018towards}, the persona \cite{chandu2019my}, the plot \cite{yao2019plan}, and the topic sequence \cite{huang2019hierarchically}.
Controllable text generation is also used to modulate the formality and politeness of emails \cite{madaan2020politeness}.
Report generation can be controlled by pulling disparate source documents into a coherent unified whole, which can use a shared set of sources such as Wikipedia article generation \cite{liu2018generating,prabhumoye-etal-2019-towards}.

Although there is a large body of prior work in controllable text generation, there is no unifying theme. 
Each work addresses a specific task in a specific context. 
In this paper we outline a new schema which connects  prior work and provides an insight into various aspects of controllable text generation. 
The schema contains five modules that cover the overall generation pipeline and provide an understanding of the effect of each component on the generation process.
Prior work has focused on specific parts of the schema that we outline here and we provide insights into their similarities.
We provide an overview of these modules and also present an exploration of the various techniques used to control and update each of these modules.

\begin{figure*}[t]
\centering
\includegraphics[scale=0.35]{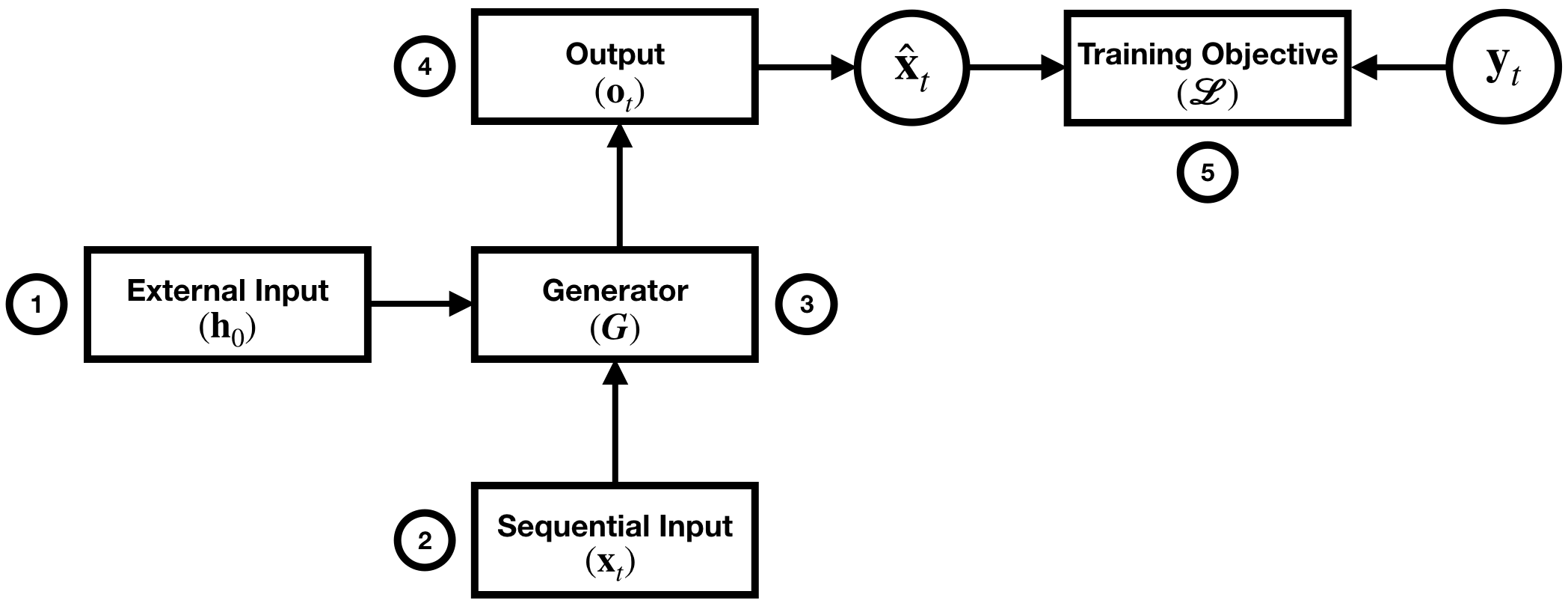}
\caption{Modules that control the generation process. Each module is numbered by the circle next to it.}
\label{fig:bg-overview}
\end{figure*}

Most of the controllable text generation tasks can be framed as conditional language generation tasks.
They have an input or a \ti{source} sequence $\B{U}$ and an output or a \ti{target} sequence $\B{Y}$ to be generated. 
In this case, we model the probability of
the \ti{target} sequence conditioned on the \ti{source} sequence given by $P(\B{Y} | \B{U}) = \prod^T_{t=1} P(\B{y}_t |\B{U}, \B{y}_{<t})$.
The generation of the target tokens of the sequence $\B{Y}$ unfolds as a time series where each token $\B{y}_t$ is generated at a time step $\B{t}$.
At a given time step $t$, a generative model takes in the previous hidden state $\B{h}_{t-1}$ and the input $\B{x}_t$ at current time step. 
It performs a set of operations denoted by $\boldsymbol{G}$ to produce the output $\B{o}_t$ which is used to predict token $\B{\hat{x}}_t$.
The ground truth token to be generated is denoted by $\B{y}_t$.

Figure~\ref{fig:bg-overview} shows the schema proposed in this work consisting of five modules which can be used for controlling the generation process: 
(1) \tb{External Input} module is responsible for the initialization $\B{h}_0$, of the generation process.
(2) \tb{Sequential Input} module is the input $\B{x}_t$ at each time step of the generation.
(3) \tb{Generator Operations} module performs consistent operations or calculations on all the input at each time step.
(4) \tb{Output} module is the output $\B{o}_t$ which is further projected on to the vocabulary space to predict the token $\B{\hat{x}}_t$ at each time step.
(5) \tb{Training Objective} module takes care of the loss functions used for training the generator.

This schema provides an insight into the contributions of the various modules for controllable text generation.
The main advantage of this schema is that it can be used with any algorithmic paradigm like sequence-to-sequence, adversarial methods, reinforcement learning, etc.
The schema can also be used with non-autoregressive algorithms which may generate text using graphical structures like trees \cite{welleck2019non,guo2019non}.
In this paper, we focus on how this schema can be used to describe controllable text generation focusing particularly on the use of autoregressive models.
This work paves way to designing new architectures based on our schema.
This can be done by identifying promising techniques for each module and then combining them.
Our schema can also be potentially used for applying these techniques on new tasks of similar nature.
It also provides a standardized framework to position and compare new architectures with existing techniques.

The prior work on unifying text generation models has mostly focused on building efficient tool-kits and modular views of generation.
For instance, \cite{reiter2000buildNLG} details seven sub-tasks which are conceptually distinct to describe the
generation process. 
These sub-tasks can be modelled separately or in some cases they may interleave. 
In \cite{reiter2000buildNLG}, these seven sub-tasks are primarily characterized as content or structure tasks.
Note that \citet{reiter2000buildNLG} is not specific to neural text generation.
Our work focuses specifically on controlling attributes in neural text generation process.
We don't divide the generation pipeline into several sub-tasks but we divide the neural text generation process into modules all of which are required for generation.
In \cite{hu2019texar}, the focus is on building a toolkit for various text generation tasks based on the three properties of versatility, modularity and extensibility.
This work enlists few model architectures and learning paradigms for various text generation tasks.
In our work, we focus only on the generation process of controllable text generation tasks.
We specifically detail the inputs, outputs and operations of the generation process.
We do not provide any specific examples of architectures but provide an overview of the basic underlying modules which can be used with any learning paradigm.
\citet{xie2017neural} provides a practical guide to the neural generation process describing it in terms of initialization, optimization, regularization and decoding strategies.
Our work on the other hand does not delve into the implementation details of the generation pipeline but provides an overall schema for understanding of the various components involved.

In the remainder of the paper, we denote the representation of the control attribute by $\B{s}$ and the representation of the input or \ti{source} sentence returned by the encoder as $\B{h}_e$.
In what follows, we first describe the possible ways of controlling attributes by modulating the \ti{external input} in \Sref{sec:bg-dec-init}, the \ti{sequential input} in \Sref{sec:bg-dec-inp}, the \ti{generator operations} in \Sref{sec:bg-gen}, the \ti{output} in \Sref{sec:bg-out} and the \ti{training objective} in \Sref{sec:bg-train-obj}.
At the end of each section, we provide an analysis of each of the techniques described and how they fit together.

\section{External Input}
\label{sec:bg-dec-init}
In this section we discuss the different techniques which can be used to control the generation process by updating the initialization of the generator $\B{h}_0$.
In the standard generation process, $\B{h}_0$ is equal to $\B{h}_e$.
This is marked as module (1) in Figure~\ref{fig:bg-overview}.

\subsection{Arithmetic or Linear Transform}
\label{sec:bg-dec-init-elem}
One of the easiest ways to control the generation is to concatenate a control vector $\B{s}$ to output of the encoder $\B{h}_e$.
The external input of the decoder $\B{h}_0$ will be $[\B{h}_e; \B{s}]$, where $[a;b]$ denotes concatenation.
Here, the control vector $\B{s}$ would provide the generator with a strong signal to guide the generation process.

\citet{FuTan} use this technique to control the style representation for their generator.
The encoder builds representation that is devoid of the style and only retains content.
The control vector for style is then concatenated to the encoder representation to initialize the decoder.
This technique is commonly used in \cite{ghazvininejad2018knowledge,zhou2018dataset,dinan2018wizard} to concatenate information from external sources to dialogue context to generate dialogue responses.
\citet{chandu2019my} concatenate personality representation $\Cal{P}$ derived from a separate corpus to generate visual stories.
They also experiment with a simple arithmetic operation on $\B{h}_e$ given by $\B{h}_0 = \B{h}_e - \Cal{S} + \Cal{P}$ to get the initialization of the generator (here $\Cal{S}$ denotes the average representation of the story).
They observed that while concatenation technique is better at preserving the meaning of the generated story, the arithmetic operation provides a better signal of the personality for the generation process.

\citet{hoang2016incorporating} uses both the concatenation technique as well as performs a linear transform of $\B{s}$ to obtain $\B{h}_0$ for language modelling task. 
The control vectors in this case represents meta data such as key-words, topics etc.
In case of the linear transform $\B{h}_0 = \T{tanh}(\B{W}_1 \B{h}_e + \B{W}_2 \B{s} + \B{b})$.
The paper also explores adding the control vector to the encoder representation ($\B{h}_0 = \B{h}_e + \B{s}$).

In case of addition, the resulting $\B{h}_0$ would be averaged representation of the input representation $\B{h}_e$ and $\B{s}$.
Information could be lost in this case as control is not explicit.
In case of concatenation, if the size of the control vector $\B{s}$ is too small compared to $\B{h}_e$, then $\B{s}$ can be over-shadowed by $\B{h}_e$ and the generator may not be able to pay attention to $\B{s}$.
Hence it is important to choose comparable dimensions for $\B{s}$ and $\B{h}_e$.
But this increases the size of model considerably and could be quite costly.
Linear transform avoids these issues and performs better than the other two techniques for \citet{hoang2016incorporating}.

\subsection{Stochastic Changes}
\label{bg:sec-stochastic}
\citet{kingma2013auto} introduce variational auto-encoder, where you can stochastically draw a continuous latent variable $\B{z}$ from a Gaussian distribution.
The initialization of the generator $\B{h}_0$ is based on this latent variable.
\citet{bowman-etal-2016-generating} use this concept for generating sentences from this continuous latent representation.
This process of changing the encoder state $\B{h}_e$ can only be used with Kullback-Leibler (KL) Divergence training objective described in \Sref{bg:sec-loss-kl}. 

In \cite{wang-etal-2019-topic}, Variational Auto-Encoder (VAE) is used to guide the generation process with topics of a document.
A gaussian mixture model is used to incorporate topics into latent variables.
In \cite{Xu2020unsupervisedVAE}, VAE is used to control for sentiment attribute in style transfer task by constraining the posterior mean to a learned probability simplex.

Such a design of controllable text generation works when the control attributes can be represented as latent variables for example style, topics, strategies etc.
This design is difficult to work for content grounded text generation tasks where specific information, keywords or entities have to guide the generation process.

\subsection{Decompose}
\label{bg:sec-decompose}
The encoder representation $\B{h}_e$ can be decomposed into multiple subspaces, each of which signifies a different attribute to be controlled.
\citet{liu2018learning} split the encoder representation $\B{h}_e$ into two components, one which represents the structure in the document and the other represents the semantic information.
This formulation was used by \cite{balachandran2020strucsum} for controlling structure in abstractive summarization.
This work performs the split with respect to the dimensions of $\B{h}_e$.
The method forces the first $n$ dimensions of $\B{h}_e$ to capture meaning and the latter to capture structure.
\citet{balachandran2020strucsum} also show quantitative and qualitative analysis on the types of structures of documents learnt by this technique.

\citet{romanov-etal-2019-adversarial} decompose the encoder representation $\B{h}_e$ into a form vector $\B{f}$ and a meaning vector $\B{m}$.
During the training phase, a \ti{discriminator} enforces $\B{m}$ to not carry any information about the form using an adversarial loss and a \ti{motivator} is used for a motivational loss that encourages $\B{f}$ to carry the information about the form.
The generation process can then be guided to adhere to the desired target form.
As opposed to splitting $\B{h}_e$ with respect to dimensions, this work learns subspaces $\B{W_m}$ and $\B{W_f}$ given by $\B{m} = \T{tanh}(\B{W}_m\B{h}_e + \B{b}_m)$ and $\B{f} = \T{tanh}(\B{W}_f\B{h}_e + \B{b}_f)$ respectively. 
When $\B{h}_e$ is projected on $\B{W}_m$, it yields the meaning vector $\B{m}$ and similarly when it is projected on $\B{W}_f$ it yields the form vector $\B{f}$.
This work shows qualitatively how $\B{m}$ and $\B{f}$ are learnt in the subspaces using t-SNE plots.
It also shows quantitatively the use of $\B{m}$ and $\B{f}$ in downstream paraphrase detection tasks.
This builds interpretable representations for control attributes.
Although, the effectiveness of this technique is not yet proven in the style transfer task or the abstractive summarization task.
In both the above mentioned works, the models learns interpretable representations of control attributes but were not able to beat state of the art methods in their respective tasks.
It is also worth noting that learning good decomposed vectors is especially hard when no supervision is provided on what the decomposed components are supposed to learn.

This technique works well when the representation space of the input $\B{x}$ can be decomposed into subspaces which can represent the control attributes.
This means that the input $\B{x}$ needs to contain signal of the control attributes.
It is unlikely to work when the control attributes need to be externally provided.
For example in case of content grounded generation tasks described in \cite{prabhumoye-etal-2019-towards,dinan2018wizard,zhou2018dataset}, the input may not necessarily contain the content that needs to be generated.
A separate input of the content to be generated is provided in these cases.

\subsection{External Feedback}
\label{bg:sec-inp-ext}
A regularizer is often used to control the external input $\B{h}_0$ to the generator.
In many cases, an adversarial loss to manipulate the latent space is used as an external feedback mechanism.
This essentially controls the latent space of the encoder which is eventually provided as an initialization to the generator.
In \cite{FuTan}, a multi-layer perceptron (MLP) is used for predicting the style labels from $\B{h}_0$.
Similarly, the adversarial loss is also used in \cite{wang2019controllable} to control the latent representation $\B{h}_0$ for style attributes.
In \cite{romanov-etal-2019-adversarial}, an adversarial loss is used to ensure that the meaning representation $\B{m}$ does not carry any style signals.
The adversarial loss is obtained by training a discriminator which takes as input a representation $\B{m}$ and indicates if it carries the target style signal.
Similarly, this work also employs a motivator loss which is the opposite of the adversarial loss to ensure that the style representation $\B{f}$ actually does carry the stylistic information.
\citet{john-etal-2019-disentangled} use multiple losses to control the style and content information represented in $\B{h}_0$.

The discriminator which provides external feedback has to be jointly trained with the generator.
This technique can be useful with the decompose technique to ensure that the decomposed sub-spaces represent the desired control attributes.

\section{Sequential Input}
\label{sec:bg-dec-inp}

In this section we discuss the different techniques which can be used to manipulate the sequential input $\B{x}_t$ to the decoder at each time step. 
$\B{x}_t$ here is used to denote the word embedding of the token at time step $t$. 
This is marked as position (2) in Figure~\ref{fig:bg-overview}.

\subsection{Arithmetic or Linear Transform}

Similar to changing the initialization, we can change the input to the decoder by concatenating the information at each time step with some additional control vector $\B{s}$.
Typically, teacher forcing method \cite{williams1989learning} is used to train the generator.
At time step $t$, the generator takes as input the word embedding $\B{x}_t$ of the word that was predicted at step $t-1$ and predicts the word to be generated $\B{y}_t$ at the current time step.
Note that $\B{x}_t = \B{y}_{t-1}$.
The input $\B{x}_t$ can be concatenated with $\B{s}$ at each time step to control the generation process.
Hence, $\B{\tilde{x}}_t = [\B{x}_t; \B{s}]$.

\citet{noraset2017definition}, use this technique in the task of definition modeling.
They concatenate word embedding vector $\B{s}$ of the word to be defined at each time step of the definition generation process.
Unfortunately, for this task, this technique has not proved to be effective compared to other techniques of controlling the generation.
\citet{zhou2018dataset} concatenate the hidden representation of the external source of information $\B{s}$ to each time step of dialogue response generation.
Similarly, \citet{prabhumoye-etal-2019-towards} also concatenate the hidden representation of the external source of information $\B{s}$ to each time step of Wikipedia update generation process.
This technique did not achieve impressive results in this work as well.
\citet{harrison2019maximizing} concatenate a side constraint $\B{s}$ which represents style and personality into the generation process.
For this task of generating language from meaning representations with stylistic variation, this method performed better than conditioning the encoder with side constraint in terms of BLEU metric.
\citet{chandu2019my} also concatenate the personality representation $\Cal{P}$ at each time step of the story generation process.
This is used to control the personality of the visual stories.
In addition to concatenation, this work proposes to modify the sequential input as $\B{\tilde{x}}_t = \B{x}_t - \Cal{S} + \Cal{P}$ (here $\Cal{S}$ denotes the average representation of the story and $\Cal{P}$ denotes the representation of the personality).
The latter technique is better at generating personality conditioned stories than the concatenation technique.
Neither of these techniques prove to be conclusively better than making similar changes to the external input module (\Sref{sec:bg-dec-init-elem}).
Note that in this technique, changes are made directly to the input of generation and not the context which is the case with external input.
Also, most of the prior work has focused on recurrent neural network and its variants for making such changes.
It could be interesting to see such changes made to transformers \cite{vaswani2017attention}.

\section{Generator Operations}
\label{sec:bg-gen}

This module takes in the external input $\B{h}_0$, the sequential input $\B{x}_t$ at time step $t$ and performs the same set of computations ($\boldsymbol{G}$) to return an output $\B{o}_t$.
Changes can be made to the set of operations $\boldsymbol{G}$ to include the the control vector $\B{s}$ in computing $\B{o}_t$.
This is shown as position (3) in Figure~\ref{fig:bg-overview}.

\subsection{Recurrent Neural Networks}
\label{sec:bg-gen-rnn}
Recurrent Neural Networks (RNNs) are designed to model sequential information. 
RNNs perform the same operations for every element of a sequence, with the output depending on previous computations. 
This recurrence serves as a form of memory. 
It allows contextual information to flow through the network so that relevant outputs from previous time steps can be applied to network operations at the current time step.
Theoretically, RNNs can make use of information in arbitrarily long sequences, but empirically, they are limited to looking back only a few steps.

The Long Short-Term Memory (LSTM) \cite{hochreiter1997long} units are a type of RNNs that have additional `memory cell' apart from standard units of basic RNNs.
The memory cell can maintain information in memory for long periods of time. A set of gates is used to control when information enters the memory, when it's output, and when it's forgotten. 
This architecture lets them learn longer-term dependencies.
The vanishing gradient problem of RNNs is resolved here. 
Gated Recurrent Units (GRUs) \cite{cho2014learning} are similar to LSTMs, but use a simplified structure designed to adaptively capture dependencies of different time scales. 
They also use a set of gates to control the flow of information, but they don't use separate memory cells, and they use fewer gates.

The computations of the RNN or its variants can be modified to account for the control attribute.
Additional gates can be added or the control attribute can be provided as an additional input to the standard gates of RNNs.
\citet{gan2017stylenet} propose a variant of the LSTM model, named factored LSTM, which controls style representation in image caption task.
The parameters of the LSTM module which are responsible to transform the input $\B{x}_t$ are factored into three components $\B{U}$, $\B{S}$ and $\B{V}$.
The operations of the input ($\B{i}_t$), forget ($\B{f}_t$) and output gate ($\B{o}_t$) are  given by:
\begin{eqnarray*}
    \B{i}_t &=& \T{sigmoid}(\B{U}_{ix} \B{S}_{ix} \B{V}_{ix} \B{x}_t + \B{W}_{ih} \B{h}_{t-1}) \\
    \B{f}_t &=& \T{sigmoid}(\B{U}_{fx} \B{S}_{fx} \B{V}_{fx} \B{x}_t + \B{W}_{fh} \B{h}_{t-1}) \\
    \B{o}_t &=& \T{sigmoid}(\B{U}_{ox} \B{S}_{ox} \B{V}_{ox} \B{x}_t + \B{W}_{oh} \B{h}_{t-1}) \\
    \B{\tilde{c}}_t &=& \T{tanh}(\B{U}_{cx} \B{S}_{cx} \B{V}_{cx} \B{x}_t + \B{W}_{ch} \B{h}_{t-1})
\end{eqnarray*}

Particularly, the matrix set $\{\B{S}\}$ is specific to each style in the task and is responsible to capture the underlying style features in the data.

In \cite{kiddon2016globally}, the GRU unit is modified to accommodate extra inputs - goal $\B{g}$ and agenda items $E^{new}_t$ in the recipe generation task.
The operation of the new component $\B{\tilde{h}}_t$ is given by:
\begin{eqnarray*}
    \B{\tilde{h}}_t = \T{tanh}(\B{W}_h \B{x}_t + \B{r}_t \odot \B{U}_h \B{h}_{t-1} + \B{s}_t \odot \B{Y}\B{g} + \\
    \B{q}_t \odot (\B{1}^{T}_{L} \B{Z} \B{E}^{new}_{t})^{T})
\end{eqnarray*}
where $\B{s}_t$ is a goal select gate and $\B{q}_t$ is a item select
gate.
With this modification, the generation process is controlled for the items to be generation in the recipe and the goal.

\citet{sem_cond_lstm} adapt the LSTM to control the dialogue act information in the generation process.
The operation to compute the cell value $\B{c}_t$ is given by:
\begin{equation*}
    \B{c}_t = \B{f}_t \odot \B{c}_{t-1} + \B{i}_t \odot \B{\tilde{c}}_t + \T{tanh}(\B{W}_d \B{d}_t)
\end{equation*}
The dialogue act representation $\B{d}_t$ is build using another LSTM cell.

RNNs, LSTMs and GRUs are commonly used to model controllable text generation tasks \cite{prabhumoye-etal-2019-towards,rao2018dear,see2017get,zhou2018dataset,FuTan}.
Most of these variants still have trouble remembering long sequences and are hence commonly used with attention mechanism (\Sref{sec:bg-out-att}) on the source sequence.

\subsection{Transformer}
Transformers are proposed by \cite{vaswani2017attention} and they rely on attention mechanism to draw global dependencies between input and output.
The Transformer uses stacked self-attention and point-wise, fully connected layers for both the encoder and decoder.
The encoder stacks $N$ identical layers, each of which has two sub-layers. 
The first sub-layer is a multi-head self-attention mechanism (\Sref{sec:bg-out-att}), and the second sub-layer is a positionwise fully connected feed-forward network. 
Each sub-layer uses residual connections around each of the sub-layers, followed by layer normalization.
The decoder has an additional third sub-layer, which performs multi-head
attention over the output of the encoder stack.

Since, attention mechanism is at the core of this generator, the decoder can attend over all positions of input sequence.
Computations over a sequence can be parallelized in this case and hence it is faster.
The modifications made to the computing units of RNN mentioned in \Sref{sec:bg-gen-rnn} which use parameters specific to control attributes such as style, dialog act etc. have not been explored with the transformers architecture.

\subsection{Pre-trained models}
Recently pre-trained conditional language models are used for text generation like GPT \cite{radford2018improving}, GPT2 \cite{radford2019language}, XLNet \cite{yang2019xlnet}, etc.
Several works have fine-tuned the pre-trained models for downstream controllable text generation tasks \cite{sudhakar2019transforming,dinan2018wizard,urbanek2019learning}.
The language modeling aspects of generation like fluency and grammaticality are already learnt if pre-trained models are used.

These models are hard to fine-tune for sequence-to-sequence tasks such as machine translation, abstractive summarization etc.
BART \cite{lewis2019bart} is a denoising autoencoder built with a sequence-to-sequence model and is
particularly effective when fine tuned for text
generation.
Alternatively, T5 \cite{raffel2019exploring} treats every NLP problem as a ``text-to-text" problem,
i.e. taking text as input and producing new text as output.
Hence, it can be adapted to controllable text generation tasks.
\citet{dathathri2019plug} propose a Plug and Play Language Model (PPLM) for controllable language generation.
It combines a pre-trained LM with one or more simple attribute classifiers that guide text generation without any further training of the LM.
This is similar to the classifier feedback technique described in~\Sref{bg:sec-loss-class}.
Some of the other techniques described in this paper such as stochastic changes~\Sref{bg:sec-stochastic} , external feedback~\Sref{bg:sec-inp-ext} and~\Sref{bg:sec-out-ext}, decompose~\Sref{bg:sec-decompose} etc would be hard to incorporate into pre-trained language models without modifying the model architecture or fine-tuning entailing the significant cost of retraining.


\section{Output}
\label{sec:bg-out}

In the standard generation process, $\B{o}_t$ is the output of the generator module which is projected to the vocabulary space to predict the token $\B{\hat{x}}_t$.
Here, we discuss the various techniques used to modulate the sequential output $\B{o}_t$ at each time step $t$, before projecting it to the vocabulary space.
This is marked as position (4) in Figure~\ref{fig:bg-overview}.

\subsection{Attention}
\label{sec:bg-out-att}

Attention is the most popular way of guiding the generation process.
It is typically used to guide the generation process to focus on the source sequence \cite{bahdanau2015neural}.
The attention calculating module takes as input the current hidden state $\B{h}_t$ of the generator at each time step $\B{t}$.
The aim of this module is to determine a context vector $\B{c}_t$ that captures relevant source-side information to help predict the token $\B{\hat{x}}_t$. 
In case of \ti{global attention}, all the hidden states of the encoder are considered to calculate the context vector $\B{c}_t$ \cite{luong2015effective}.
This faces the the downside of expensive calculation especially for longer source sequences like documents.
To overcome this challenge, \ti{local attention} only chooses to focus only on a small subset of the source positions per target word.
In this case, $\B{c}_t$ is calculated over a window of size $D$ of the source hidden states.

\citet{vaswani2017attention} view attention as a mapping a query and a set of key-value pairs to an output, where the query, keys, values, and output are all vectors. 
The output is computed as a weighted sum of the values, where the weight assigned to each value is computed by a compatibility function of the query with the corresponding key.
This work proposes the simultaneous use of \ti{scaled dot-product} attention which helps in parallelizing computation and a \ti{multi-headed} attention which allows the model to jointly attend to information from different representation subspaces at different positions.

\citet{sudhakar2019transforming} use self-attention to control for style by simply adding a special target style token in the source sequence. 
\citet{dinan2018wizard} also use transformers to attend over information from external document for guided dialogue response generation.
\cite{zhang2018personalizing} uses the encoded representation of personas to compute the attention weights $\B{a}_t$ at a given time step of the decoder.
The attention is re-weighted according to the persona of the response to be generated in dialogue.
So far, work has not been done to modulate the attention weights to control for attributes like style, topic, content etc.

\subsection{External Feedback}
\label{bg:sec-out-ext}
The output latent space of the generator can be controlled by external feedback.
Similar to changing the external input $\B{h}_0$, the output latent space can also be changed using adversarial loss.
In \cite{logeswaran2018content}, an adversarial loss is used which encourages the generation realistic and attribute compatible sentences. 
The adversarial loss tries to match the distribution of sentence and attribute vector pairs $(\B{x}, \B{s})$ where the sentence can either be a real or generated sentence.
Similarly, in \cite{shen2017style}, a two discriminator losses in the style transfer task.
Each discriminator is trained to distinguish between a sentence which came from the real target attribute distribution and a sentence that was transferred from source to target attribute.
This work uses Professor-Forcing~\cite{lamb2016professor} to match the hidden states of the generator and the discriminator.
\citet{gong-etal-2019-reinforcement} also control the output latent space by providing different types of rewards like style reward, semantic reward and fluency reward in the reinforcement learning setup.
The discriminator used to obtain the adversarial loss has to be jointly trained with the generator.

\subsection{Arithmetic or Linear Transform}
\citet{hoang2016incorporating} demonstrate three simple ways of changing the output $\B{o}_t$ of an RNN to control for meta information like topic, keywords etc.
The three ways demonstrated in \cite{hoang2016incorporating} are: (1) addition, where the modified output $\B{\tilde{o}}_t$ is $\B{\tilde{o}}_t = \B{o}_t + \B{s}$, 
(2) concatenation, where the modified output $\B{o}_t$ ($\B{\tilde{o}}_t = [\B{o}_t; \B{s}]$), and
(3) using a perceptron layer dependent on $\B{s}$ and $\B{o}_t$. 
In this case, $\B{\tilde{o}}_t$ is given by $\B{\tilde{o}}_t = \T{tanh}(\B{W}_o \B{o}_t + \B{W}_s \B{s} + \B{b}_o)$.
In each of the three cases, the modified output $\B{\tilde{o}}_t$ is then projected to the vocabulary space to predict the token $\B{\hat{x}}_t$.

\section{Training Objective}
\label{sec:bg-train-obj}

In this section we describe various methods used to control the generation using objective functions.
The output $\B{o}_t$ at each time step $t$ of the generation process is projected to the vocabulary space using a linear transform ($\B{\hat{o}}_t = \B{W}_o \B{o}_t + \B{b}$).
A token $\B{\hat{x}}_t$ is predicted from the vocabulary by passing $\B{\hat{o}}_t$ through a softmax function and taking the max value.
The predicted token $\B{\hat{x}}_t$ is compared with the reference token $\B{y}_t$ using a loss function.
This loss function can be tweaked to ensure that the generated text carries the desired control attributes.

\subsection{General Loss Objective}
Here, we describe the loss objectives commonly used in natural language generation tasks. 
These loss objectives do not try to control for any attribute. 
Instead they try to ensure fluent, grammatical and diverse generations.

\paragraph{Cross Entropy Loss:}
This is the basic loss used to compare the generated tokens with the reference tokens and is used in all text generation process.
At each time step $t$, the generation has to predict a token from the vocabulary.
Hence, it could be seen as a classification problem with number of classes being equal to vocabulary size. 
The categorical cross entropy loss is given by $- \Sigma^{M}_{c=1} \B{y}_{t, c} \T{log} (p_{t, c})$.
Here $p_{t, c}$ is the probability of the token $c$ at time step $t$. 
Note that $p_t = \T{softmax} (\B{\tilde{o}}_t)$ is the probability distribution over the vocabulary.

\paragraph{Unlikelihood loss: } This maintains a set of negative candidates which is based on repeating tokens or n-grams and frequent tokens~\cite{Welleck2020Neural}.
This set is updated at each time step as tokens are generated.
This works at both token and sequence level and the objective tries to minimize the repetitions in generations.
This is used at train time in augmentation with the maximum likelihood objective and can be used for any task.

\paragraph{Decoding strategies: } These strategies are not used as a loss objective during training.
Many of these objectives rely on post-hoc decoding strategies such as stochastic decoding which include Top $k$-sampling \cite{fan:2018}, nucleus sampling \cite{Holtzman2020The}, or beam search variants \cite{paulus2018a,kulikov-etal-2019-importance,vijayakumar2018diverse,holtzman2018learning}.

Specifically, we discuss the Diversity-Promoting objective which is used to generate a varied set of sentences given similar inputs.
Particularly, \citet{li:2015} use Maximum Mutual Information (MMI) as an objective function for the dialogue response generation task.
Most generation systems use maximum likelihood objective but this objective additionally tries to reduce the proportion of generic responses.
It is given by:
\begin{equation*}
    \B{\hat{T}} = \T{argmax}_{T} \{ \T{log} p(\B{T} | \B{S}) - \lambda \T{log} p(\B{T})\}
\end{equation*}
where $\B{\hat{T}}$ is the generated target sequence, $\B{T}$ is the reference target sequence and $\B{S}$ is the source sequence. 
The second term controls the generation of the high frequency or the generic target sequences.
Note that this objective is only used during the inference and the generators are trained using cross entropy loss.
\citet{zhang2018personalizing}, also use a diversity encouraging objective for dialogue response generation.
They train a discriminator to calculate similarity between the source $\B{S}$ and target $\B{T}$ ($D_{\psi}(\B{T}, \B{S})$) , as well as between the source $\B{S}$ and the generated target $\B{\hat{T}}$ ($D_{\psi}(\B{\hat{T}}, \B{S})$).
They finally try to minimize the difference between $D_{\psi}(\B{T}, \B{S})$ and $D_{\psi}(\B{\hat{T}}, \B{S})$.

\subsection{KL Divergence}
\label{bg:sec-loss-kl}
The Kullback-Leibler (KL) Divergence score, quantifies how much one probability distribution differs from another probability distribution.
The KL divergence between two distributions $\B{\Cal{Q}}$ and $\B{\Cal{P}}$ is often stated using the notation $\T{KL} (\B{\Cal{P}} \parallel \B{\Cal{Q}})$,
where the  operator ``$\parallel$'' indicates \ti{divergence} or $\B{\Cal{P}}$'s divergence from $\B{\Cal{Q}}$.
Note that KL Divergence is not symmetric i.e $\T{KL} (\B{\Cal{P}} \parallel \B{\Cal{Q}}) \neq \T{KL} (\B{\Cal{Q}} \parallel \B{\Cal{P}})$.
KL divergence can be used to minimize the information loss while approximating a distribution. 
In text generation, the KL Divergence is combined with the evidence lower bound (ELBO) to approximately maximize the marginal likelihood of data $p(\B{x})$ which helps in better generations.
This objective is used in variational autoencoders and its variants in combination with sampling techniques described in \Sref{bg:sec-stochastic}.
This objective fits in the controllable text generation paradigm because it allows you to approximate the posterior distribution of the control variables in the latent $\B{z}$-space.

\subsection{Classifier Loss}
\label{bg:sec-loss-class}

This loss is specifically used to ensure that the generated tokens $\B{\hat{x}}$ comply with the control attributes $\B{s}$.
Note the difference between this loss and the external feedback loss used for the \ti{external input} module and the \ti{output} module is that this loss operates at the token level and the external feedback loss works on the latent hidden representations.

In case of style transfer task, this loss is used to guide the generation process to output the target style tokens.
Some works \cite{prabhumoye:2018,sudhakar2019transforming,hu2017toward} use this loss to discriminate between all the styles in their task (one verses all fashion). 
This type of design will suffer from low accuracy scores when the number of styles increases.
To counter this problem, this loss can be setup  to calculate if the generated sentence $\B{\hat{x}}$ belongs to style $\B{s_1}$ or not and similarly to calculate another separate loss term for each style \cite{chandu2019my}.
This type of loss design encounters increasing number of loss terms depending on the number of styles.
The third way to motivate this loss term is to discriminating between a sentence $\B{x}$ from data which belongs to style $\B{s_1}$ and a generated sentence $\B{\hat{x}}$ which belongs to the same style $\B{s_1}$ \cite{yang2018unsupervised}.
Again, you would need as many loss terms as the number of styles in this case.
All of these works use cross entropy loss function to measure their losses.

\citet{hu2019makes} use a classifier based loss in the visual storytelling task.
The classifier is a pre-trained language model \cite{devlin2019bert} used to measure the coherence between generated sentences of the story.
Particularly, the classifier takes as input two sentences at a time $\B{\hat{x}}_1$ and $\B{\hat{x}}_2$ and outputs a binary label which indicates if $\B{\hat{x}}_2$ follows $\B{\hat{x}}_1$.
In this case, the control variable is coherence in stories which is used to guide the generator to produce consistent sentences.

\subsection{Task Specific Loss}
Depending on the end task and the attribute to be controlled, you can design different loss objectives to ensure that generations abide by the target attributes.

\paragraph{Strategy Loss:} \citet{Zhou2020Augmenting} use a dialogue strategy based objective to generate responses for negotiation tasks.
This task has ground truth strategies that lead to better negotiations.
This loss captures the probability of a particular strategy occurring for the next utterance given the dialogue history. 
It guides the generator to align the responses with particular strategies.

\paragraph{Coverage Loss:}
Generating repeated words or phrases is a common problem for text generation systems, and this becomes especially pronounced for multi-sentence text generation task such as abstractive document summarization.
\citet{see2017get} introduce a \ti{coverage loss} which penalizes repeatedly attending to the same locations of the source document.

\paragraph{Structure loss:}
\citet{li-etal-2018-improving-neural} introduce two new loss objectives \ti{structural compression} and \ti{structural coverage} based on sentence-level attention.
These objectives are specially designed for the task of abstractive document summarization.
\ti{structural compression} is used to generate a sentence by compressing several specific source sentences and \ti{structural coverage} is used to cover more salient information of the original document.
These objectives leverage document structure in document summarization, and explore the effectiveness of capturing structural properties of document summarization by regularization of the generative model to generate more informative and concise summaries.


\section{Discussion}
\label{sec:bg-discussion}

\paragraph{Discrete space issues: } The classifier loss (\Sref{bg:sec-loss-class}) is used to determine if the generated tokens $\B{\hat{x}}$ are in accordance with the target control attribute $\B{s}$.
To calculate the loss, the generated tokens $\B{\hat{x}}$ are provided as input to the classifier.
If the tokens in this case are generated using the \emph{argmax} then this function is not differentiable.
Hence, passing tokens effectively to the classifier is a challenge.

In~\cite{yu2017seqgan}, the REINFORCE~\cite{williams1992simple} algorithm is used and rewards are calculated using Monte Carlo search sampling for the next tokens.
This technique is known to be unstable due to the high variance of the sampled gradient during training~\cite{shen2017style}.
\citet{kusner2016gans} introduce the Gumbel-softmax distribution as a solution.
It approximates the multinomial distribution parameterized in terms of the softmax distribution.
Here the predicted token is:
\begin{equation*}
    \B{\hat{x}}_t = \texttt{softmax}(1/\mathbf{\tau}(\B{\hat{o}}_t + \B{g}_t)),
\end{equation*}
where $\B{\hat{o}}_t$ is described in (\Sref{sec:bg-train-obj}), $\mathbf{\tau}$ is temperature parameter and $\B{g}_t$ is sampled independently from the Gumbel distribution.
\citet{hu2017toward} use this technique without sampling from the Gumbel distribution but by only training the temperature parameter.

\paragraph{Combined module architecture: }
It is also possible to combine techniques from multiple modules to control the generation process.
We mention some of the prior works that have successfully combined various modules here.
\citet{hu2017toward} combine stochastic changes (\Sref{bg:sec-stochastic}), KL Divergence loss (\Sref{bg:sec-loss-kl}) and a classifier loss (\Sref{bg:sec-loss-class}).
It adopts a variational auto-encoder along with KL divergence loss objective and further adds a discriminator loss which signifies if the generated sentence belong to the target attribute.
As mentioned earlier, \citet{romanov-etal-2019-adversarial} combine the decomposition of the external input (\Sref{bg:sec-decompose}) with external feedback provided to the external input (\Sref{bg:sec-inp-ext}).
External feedback is used to ensure that the decomposed latent sub-spaces represent the desired target attributes. 
\citet{hu2018on} establishes formal connections between generative adversarial networks (related to \Sref{bg:sec-out-ext} and \Sref{bg:sec-loss-class}) and  variational auto-encoders (related to \Sref{bg:sec-stochastic} and \Sref{bg:sec-loss-kl}).
Determining the best possible combination of modules through empirical evaluation remains an open challenge.

\section{Conclusion and Future Work}
In this paper we propose a new schema to organize the prior work in controllable text generation.
The schema contains five modules, each of which plays an important role in the generation process.
We detail the various techniques used to modulate each of the five modules to perform controllable text generation.
We also provide theoretical understanding and qualitative analysis of these techniques.
This understanding paves way to new architectures based on combinations of these modules.
The future work will focus on empirical comparison of these techniques to gain an insight into their usefulness and strength.

\section*{Acknowledgments}
This work was supported in part by NSF IIS1763562, and ONR Grant N000141812861. 
We would like to thank Elijah Mayfield, Sai Krishna Rallabandi, Shruti Palaskar, Aman Madaan, Bhuwan Dhingra, Harsh Jhamtani and Khyathi Chandu for valuable discussions at earlier
stages of this work.
We are also grateful to the anonymous reviewers for their constructive feedback.

\bibliography{acl2020}
\bibliographystyle{acl_natbib}

\appendix



\end{document}